\documentclass[10pt,twocolumn,letterpaper]{article}
\usepackage{graphicx}
\usepackage{rotating}
\usepackage{booktabs}
\usepackage{multirow}
\usepackage{caption}
\usepackage{subcaption} 

\newcommand{\qualdir}{images/qualitative_results}

\newcommand{\colimg}[1]{\includegraphics[width=0.13\linewidth]{#1}}

\newcommand{\qualrow}[1]{%
  \colimg{\qualdir/anomalous/#1.png} &
  \colimg{\qualdir/ours/#1.png} &
  \colimg{\qualdir/ours/#1binary.png} &
  \colimg{\qualdir/GLCM/#1_glcm_energy_heatmap.png} &
  \colimg{\qualdir/GLSR/#1.png} &
  \colimg{\qualdir/SR/#1_saliency.png} &
  \colimg{\qualdir/WinCLIP/#1_heatmap.jpg} \\
}

\usepackage{wacv}              

\usepackage{graphicx}
\usepackage{amsmath}
\usepackage{mathtools}
\usepackage{amssymb}
\usepackage{booktabs}

\usepackage[pagebackref,breaklinks,colorlinks]{hyperref}

\usepackage[capitalize]{cleveref}
\crefname{section}{Sec.}{Secs.}
\Crefname{section}{Section}{Sections}
\Crefname{table}{Table}{Tables}
\crefname{table}{Tab.}{Tabs.}

\def\wacvPaperID{2729} 
\def\confName{WACV}
\def\confYear{2025}

\begin{document}

\title{A Single Image Is All You Need: Zero-Shot Anomaly Localization Without Training Data\thanks{Preprint. Under review.}}

\author{Mehrdad Moradi$^{1}$ \quad
Shengzhe Chen$^{2}$ \quad
Hao Yan$^{2}$ \quad
Kamran Paynabar$^{1}$ \\
{\tt\small mmoradi6@gatech.edu \quad schen415@asu.edu \quad haoyan@asu.edu \quad kamran.paynabar@isye.gatech.edu}
\\
$^{1}$Georgia Tech \\
$^{2}$Arizona State University \\
}
\maketitle

\begin{abstract}
Anomaly detection in images is typically addressed by learning from collections of training data or relying on reference samples. In many real-world scenarios, however, such training data may be unavailable, and only the test image itself is provided. We address this zero-shot setting by proposing a single-image anomaly localization method that leverages the inductive bias of convolutional neural networks, inspired by Deep Image Prior (DIP). Our method is named Single Shot Decomposition Network (SSDnet). Our key assumption is that natural images often exhibit unified textures and patterns, and that anomalies manifest as localized deviations from these repetitive or stochastic patterns. To learn the deep image prior, we design a patch-based training framework where the input image is fed directly into the network for self-reconstruction, rather than mapping random noise to the image as done in DIP. To avoid the model simply learning an identity mapping, we apply masking, patch shuffling, and small Gaussian noise. In addition, we use a perceptual loss based on inner-product similarity to capture structure beyond pixel fidelity. Our approach needs no external training data, labels, or references, and remains robust in the presence of noise or missing pixels. SSDnet achieves 0.99 AUROC and 0.60 AUPRC on MVTec-AD and 0.98 AUROC and 0.67 AUPRC on the fabric dataset, outperforming state-of-the-art methods. The implementation code will be released at \href{https://github.com/mehrdadmoradi124/SSDnet}{https://github.com/mehrdadmoradi124/SSDnet}
\end{abstract}

\section{Introduction}
\label{sec:intro}
Zero-shot anomaly detection (ZSAD) has gained significant interest recently. Typically, ZSAD assumes that a model has been trained on training data and at the test time, it should detect anomalies on test data drawn from a distribution different from the training data. The main goal in ZSAD is therefore to train models that can generalize to unseen data. However, when no training data is available, ZSAD becomes very challenging. This is a realistic setting when data collection is costly or the number of data points is very small in materials science, or additive manufacturing. 

To perform ZSAD without training data, the normal pattern must be learned directly from the test image, and anomalous pixels are identified by their deviations from this pattern. Classical statistical descriptors have been employed for this purpose. For example, gray level co-occurrence matrices (GLCM) \cite{haralick1973textural-GLCM_orig}\cite{okarma2016noref_GLCM} \cite{raheja2013real_GLCM} capture the joint distribution of intensity pairs at specific offsets in distance and angle. Other descriptors include local and global entropy \cite{fastowicz2019objective_entropy}, structural similarity (SSIM)  \cite{okarma2019adaptation_SSIM} and Hough transform for average line length\cite{fastowicz2019quality_HOUGH}. However, such descriptors often fail when images exhibit complex or stochastic textures, as they cannot capture the shared structural patterns across regions of the image.

A more powerful alternative to simple descriptors is low-rank decomposition, where the normal component is assumed to exhibit a low-rank structure while anomalies are sparse. For instance, PG-LSR \cite{cao2017fabric_PGLSR} decomposes an image into low-rank and anomaly components by minimizing a Frobenius-norm objective, guided by a coarse anomaly map derived from precomputed texture features. Robust PCA (RPCA) \cite{candes2011robust_RPCA} enforces low-rankness via a nuclear norm penalty and sparsity via an $\ell_1$ norm. Similarly, SSD \cite{yan2017anomaly_SSD} models the normal component with B-spline bases while regularizing the difference between neighboring coefficients with an $\ell_2$ norm to enforce smoothness in the normal background.  Despite their effectiveness, these methods rely heavily on the assumption that normal and anomalous components are strictly low-rank and sparse. This assumption breaks down for data with nonlinear or highly complex patterns, causing low-rank methods to fail in capturing the underlying normal structure.

To overcome the limitations of low-rank methods, researchers have turned to foundation models for zero-shot anomaly detection. WinCLIP \cite{jeong2023winclip} applied CLIP \cite{radford2021learning_CLIP} in a sliding-window fashion, computing similarities between image patch embeddings and text prompts corresponding to normal and anomalous classes. While effective without additional training, this approach faces several fundamental limitations. First, prompt design must be carefully tailored to the specific object and anomaly type, limiting generality. Second, window-based detection cannot localize fine-grained pixel-level anomalies. Third, the method relies solely on CLIP’s pretrained knowledge and does not adapt to the unique characteristics of each test image, which may deviate from the training distribution.

To overcome the restrictive assumptions of low-rank decomposition methods and zero-shot foundation models, we propose Single Shot Decomposition Network (SSDnet). Our method trains a neural network on overlapping patches of a single image, leveraging the inductive bias of convolutional neural networks \cite{ulyanov2018deep_DIP}\cite{gandelsman2019double_DIP}\cite{arora2019implicit_bias_deep_matrix} to capture the underlying patterns of the data. To model complex textures such as stochastic fabrics, we introduce a perceptual loss based on inner-product similarity of embeddings.
Our main contributions are:
\begin{itemize}
\item We demonstrate that accurate anomaly localization from a single image is feasible and robust to noise and missing pixels.
\item We design an optimization framework that captures shared structures without assumptions on data distribution, enhanced with a perceptual loss to effectively model complex patterns.
\item We provide a flexible formulation that allows practitioners to adjust resolutions and aggregation functions across domains.
\end{itemize}

\section{Related Work}
\label{sec:related}
\subsection{Zero-Shot Anomaly Detection}
Recently, CLIP \cite{radford2021learning_CLIP} has emerged as a prominent tool for zero-shot anomaly detection. WinCLIP \cite{jeong2023winclip} applies sliding windows of varying sizes across the image and computes the similarity score between each window and compositional prompt ensembles. These ensembles are template prompts that describe anomalous or normal states, optionally incorporating domain-specific knowledge of the test image. The final anomaly map is obtained by aggregating similarity maps across multiple resolutions.

An important limitation of this approach lies in designing prompt templates that effectively capture image-specific anomalies. To address this, APRIL-GAN \cite{chen2023april_GAN_CLIP} extended CLIP by inserting fully connected layers into the image encoder, mapping features to the shared embedding space. These layers were trained with focal and dice losses on a training set while the original CLIP weights remained frozen. Several studies have proposed learning text prompts via a segmentation loss function \cite{zhou2023anomalyclip, cheng2025efficient_CLIP_MEASUREMENT, gu2024filo}. In particular, MGVCLIP \cite{cheng2025efficient_CLIP_MEASUREMENT} introduced lightweight convolutional layers after each layer of the vision transformer while freezing the original CLIP encoder weights. They leveraged multi-layer image features and projected them to the text-embedding space using a learned fully connected layer, enabling the text prompts to be optimized jointly with the segmentation objective. A key limitation of WinCLIP is its tendency to emphasize object class rather than anomaly state. To address this, AnomalyCLIP \cite{zhou2023anomalyclip} learns class-agnostic prompts using diagonal attention and by injecting learnable tokens into the middle layers of the text encoder. More recently, Bayes-PFL \cite{qu2025bayesian_CLIP} proposed a Bayesian prompt bank that models two prompt distributions: one image-specific and the other image-agnostic.

Beyond text prompts, C2AD \cite{10888992_c2ad_CLIP} introduces a guided visual prompt to enhance semantic and contextual consistency. The method treats the original CLIP model (without the visual prompt) as a teacher and the prompted model as a student, enforcing the student’s embeddings to align with the teacher’s. Additionally, it matches the correlation structure of a batch of image embeddings with that of the teacher. To further ensure contextual consistency, the same anomaly objects are enforced to have the same embeddings in different contexts. The new contexts were generated using Inpaint Anything \cite{yu2023inpaint_anything}, Segment Anything \cite{kirillov2023_SAM}, and diffusion models \cite{ho2020ddpm}. In a related direction, \cite{luddecke2022image_CLIP_UNET} proposed CLIPSeg, which augments CLIP with a transformer-based decoder connected to the image encoder through UNet-style skip connections. They project an interpolated conditional vector derived from both image and text embeddings to the image decoder that generates the segmentation mask.

Beyond CLIP-based approaches, other ZSAD methods leverage alternative foundation models. Hou et al. \cite{hou2024enhancing_CASCADED_SAM_CLIP} first identify the top-K anomaly candidates using CLIP and then apply SAM in a cascaded manner to refine the anomaly bounding boxes. FiLo \cite{gu2024filo} incorporate Grounding DINO \cite{liu2024grounding_DINO} to filter and refine anomaly maps initially generated by CLIP, improving localization accuracy.

These approaches depend on large-scale foundation models and typically require additional training on data subsets to optimize prompts or auxiliary weights. In contrast, our method is training-free, lightweight, and operates without reliance on external datasets.

\subsection{Matrix Decomposition Models}
For general anomaly detection, \cite{candes2011robust_RPCA} introduced Robust Principal Component Analysis (RPCA), which formulates anomaly detection as a principal component pursuit problem by decomposing data into a low-rank normal component and a sparse anomaly component. Low-rankness is enforced via the nuclear norm, while sparsity is enforced via the $\ell_1$-norm. When applying RPCA to a single image, the image must be divided into patches, with the assumption that the underlying normal pattern is low-rank. If the pattern is instead smooth, one can apply Smooth Sparse Decomposition (SSD) directly to the full image. \cite{yan2017anomaly_SSD} proposed SSD, which decomposes an image into smooth and sparse components using basis functions (e.g., B-splines). Smoothness is enforced by penalizing differences among neighboring B-spline coefficients, while sparsity is enforced by an $\ell_1$-penalty on the anomaly coefficients.

Instead of decomposing an image into only normal and anomaly components, some works introduce a third noise component to capture spurious defects caused by illumination variations and shadows \cite{shi2020fabric_gradient_graph,9123438_WDLRD,8826547_GLNR}. In this framework, \cite{shi2020fabric_gradient_graph} and GLNR \cite{8826547_GLNR} regularize the noise term with a Frobenius norm penalty, while WDLRD \cite{9123438_WDLRD} enforces sparsity on the noise component via an $\ell_1$-norm.

Some works reformulate decomposition as a two-stage process \cite{shi2020fabric_gradient_graph,9123438_WDLRD,8826547_GLNR,cao2017fabric_PGLSR,8402199_WLRR}. In the first stage, a coarse anomaly map is generated and used as a guiding matrix. In the second stage, this guiding matrix constrains the decomposition by encouraging the anomaly component to align with the coarse map. PG-LSR \cite{cao2017fabric_PGLSR}, WDLRD \cite{9123438_WDLRD}, and W-LRR \cite{8402199_WLRR} derive the guiding matrix from precomputed texture features, measuring the distance between each patch’s features and those of other patches—the larger the distance, the more likely the patch is anomalous. By contrast, GLNR \cite{8826547_GLNR} constructs the guiding matrix from gradient information, under the assumption that anomalies predominantly occur along edges.

SSDnet generalizes low-rank decomposition methods by removing restrictive assumptions of sparsity or low-rankness on anomaly and normal components. Instead, it assumes that the dominant structure of the image is normal and leverages the inductive bias of convolutional neural networks to capture this shared pattern. Unlike prior approaches, SSDnet operates in zero-shot manner, and is specifically designed for the single-image anomaly detection setting with no training data.
\section{Methodology}
\label{sec:method}
\subsection{Method Overview}
Given a single image $y$, our goal is to decompose it into normal, anomalous, and noise components:
\begin{equation}
    y = \mu + a + \epsilon,
\end{equation}
where $\mu$ denotes the normal component, $a$ the anomalous component, and $\epsilon$ the residual noise.  

\begin{figure*}[t]
    \centering
    \includegraphics[width=\linewidth]{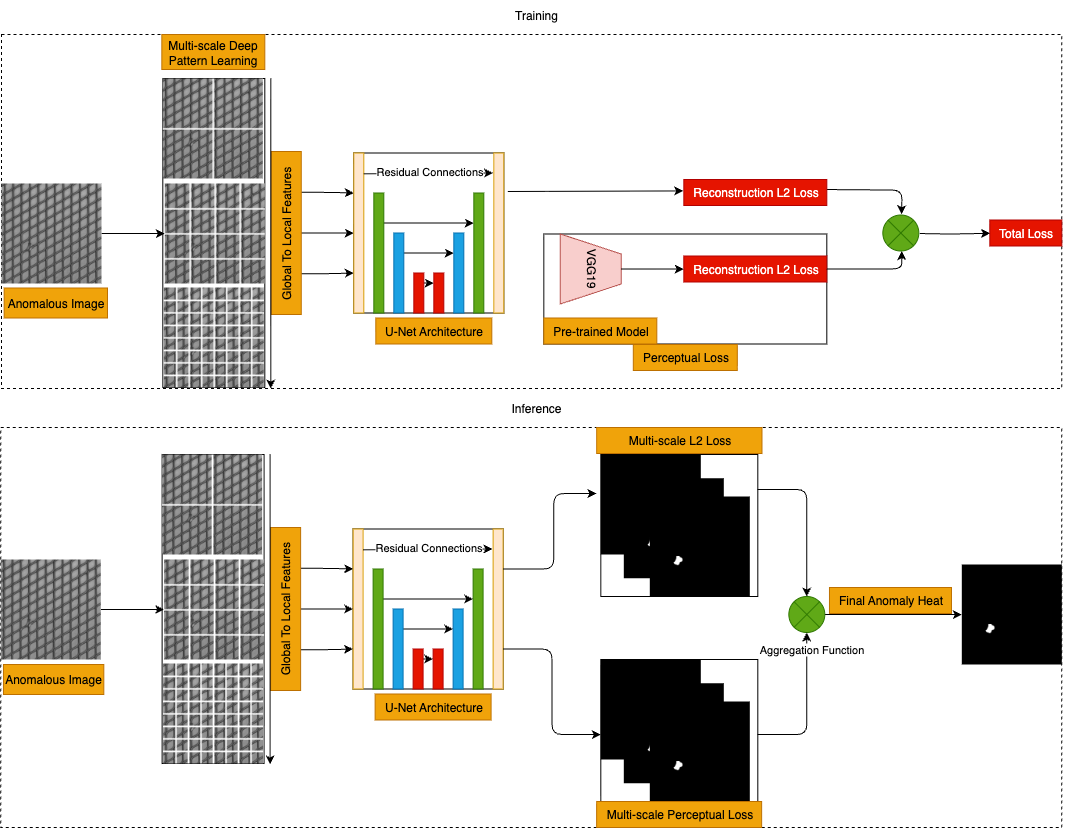}
    \caption{SSDnet overview.}
    \label{fig:method_overview}
\end{figure*}

We model the normal component as the output of a neural network $f_\theta(y)$, leveraging the inductive bias of convolutional architectures. The anomalous component is defined implicitly via a loss function $L(y, f_\theta(y))$. To ensure that $f_\theta$ captures the underlying normal pattern, we divide $y$ into overlapping patches of varying sizes and train the network to minimize the following objective:
\begin{align}
\label{eq:optimization}
&\min_{\theta} \; \mathcal{L}(y, f_\theta) \\
&= \Lambda_{r\in R} \;\frac{1}{N_r}\sum_{p \in \mathcal{P}_r} 
\Bigg( 
w_{rec}\,\big\| R_{p,r} \odot y - R_{p,r} \odot f_\theta(y)\big\|_2^2 \notag \\
& \qquad - w_{perc}\,\big\langle \phi(R_{p,r}\odot y), \phi(R_{p,r} \odot f_\theta(y)) \big\rangle
\Bigg) + \mathcal{R}(\theta)\notag
\end{align}
Here, $R_{p,r} \in \{0,1\}^{H \times W}$ is a binary masking matrix of the same size as the image 
$y \in \mathbb{R}^{H \times W}$. It is zero everywhere except on the patch of size $m_r \times m_r$ 
at position $p=(p_x,p_y)$, where
\[
(R_{p,r})_{i,j} = 
\begin{cases}
1 & \text{if } p_x \leq i < p_x+m_r,\;\; p_y \leq j < p_y+m_r, \\
0 & \text{otherwise}.
\end{cases}
\]
Applying $R_{p,r}$ with the Hadamard product $\odot$ extracts the patch region from $y$. 
In Equation~\ref{eq:optimization}, $\phi(\cdot)$ denotes the embedding function for perceptual 
similarity, and $\mathcal{R}(\theta)$ is a regularization term. The operator $\Lambda$ aggregates 
losses across resolutions, $\mathcal{P}_r$ is the set of patch indices at resolution $r$, and 
$N_r = |\mathcal{P}_r|$ is the number of patches, used to normalize each resolution so that no 
single resolution dominates the optimization. Finally, $R$ is the set of all resolutions, and 
$w_{\text{rec}}$ and $w_{\text{perc}}$ are the reconstruction and perceptual loss weights. An overview of the method in training and inference is shown in Figure \ref{fig:method_overview}.
\subsection{Neural Network Architecture as Regularizer}

The architecture of the neural network can act as a regularizer. Convolutional neural networks (CNNs), in particular, are known to embed strong inductive biases about natural images. Ulyanov et al. \cite{ulyanov2018deep_DIP} showed that an untrained encoder–decoder CNN with residual connections can serve as a ``deep image prior,'' capable of solving tasks such as denoising, inpainting, and artifact removal by simply optimizing it to map random noise to a single image. Gandelsman et al. \cite{gandelsman2019double_DIP} further extended this idea by turning vision tasks such as segmentation, dehazing, and transparency separation into image decomposition to two such priors.

Beyond image restoration, the implicit regularization effect of overparameterized neural networks has been studied in broader contexts. Saragadam et al. \cite{Saragadam2024_DIP_tensor_decomposition} demonstrated that residual CNNs provide beneficial inductive biases for matrix and tensor factorization. Similarly, Arora et al. \cite{arora2019implicit_bias_deep_matrix} analyzed deep linear networks for matrix completion and sensing, showing that increased network depth biases the solution toward low-rank structures and improves recovery. These works highlight the implicit regularization properties of CNNs.

For anomaly detection, encoder–decoder models provide an additional advantage. By projecting high-dimensional images onto lower-dimensional manifolds, such architectures naturally reconstruct common structures more easily than rare or anomalous pixels. Zhou et al. \cite{zhou2017anomaly_robust_deep_AE} leveraged this property in robust autoencoders, where normal patterns were faithfully reconstructed while anomalies were separated by a sparse component. This property makes encoder–decoders particularly well-suited for separating normal from anomalous content in single-image settings.

\begin{figure*}[t]
    \centering
    \includegraphics[width=\linewidth]{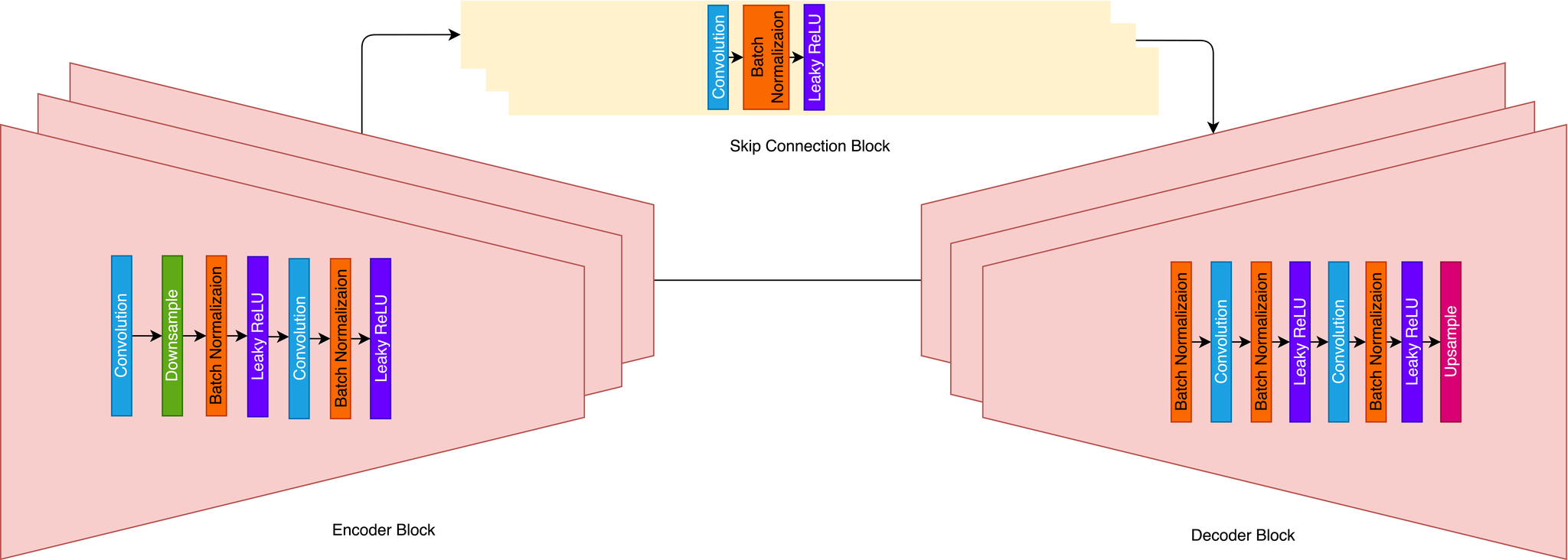}
    \caption{Neural network architecture used in SSDnet, adapted from DIP \cite{ulyanov2018deep_DIP}.}
    \label{fig:architecture}
\end{figure*}

Motivated by these observations, SSDnet adopts the encoder–decoder with residual connections originally proposed in \cite{ulyanov2018deep_DIP}. In our framework, this architectural choice acts as the regularizer $\mathcal{R}(\theta)$ in Equation \ref{eq:optimization}, enforcing the model to capture shared patterns within an image while suppressing anomalies. An overview of the architecture is shown in Figure \ref{fig:architecture}, and implementation details will be provided in our released code.

\subsection{Perceptual Loss Function}
Perceptual loss is defined on feature maps extracted from different layers of a pretrained network such as VGG \cite{simonyan2014very_VGG}. VGG consists of convolutional, pooling, and fully connected layers, and is pretrained on ImageNet \cite{russakovsky2015imagenet} for image classification. \cite{gatys2015neural_perceptual1} introduced Euclidean distance on both feature maps and their Gram matrices to capture image style and content. \cite{gatys2015texture_perceptual2} applied Gram-matrix reconstruction loss for texture synthesis, while \cite{johnson2016perceptual3} used similar losses for style transfer and super-resolution. 

Beyond Euclidean distance, \cite{zhang2018super_identity} proposed normalized Euclidean distance between embeddings, while \cite{patil2025motionswap} employed cosine similarity as the perceptual loss. In the Appendix \ref{app:perceptual}, we show that cosine similarity is equivalent to normalized Euclidean distance.

In our method, we use feature maps from the eighth layer of VGG19. Input images are resized to $224 \times 224$ RGB and yield feature maps of dimension $(128,112,112)$. Unlike prior works relying on reconstruction or Gram-matrix losses, we minimize the negative inner-product, effectively maximizing similarity between feature maps. Unlike cosine similarity–based perceptual losses \cite{patil2025motionswap,zhang2018super_identity}, which encourage only angular alignment between feature embeddings, our inner-product formulation enforces both directional alignment and norm preservation, which encourages the strength of aligned features.
\subsection{Optimization Design Choices}
Our model provides substantial flexibility, allowing practitioners to tailor the optimization in Equation \ref{eq:optimization} to their specific needs. The choice of resolutions $m_r$ can be guided by prior knowledge of anomaly sizes—when such information is available, selecting an appropriate resolution in advance is straightforward. In cases where no prior information is accessible, one can instead employ a wide range of resolutions to ensure coverage across potential anomaly scales.

For the aggregation function $\Lambda$, different strategies can be employed. A straightforward choice is the $\max$ operator, where the anomaly score of a pixel is determined by its highest score across resolutions. This makes the model more sensitive, as a pixel flagged anomalous in any resolution is treated as anomalous overall. A more conservative alternative is the $\min$ operator, which requires a pixel to be consistently detected as anomalous across all resolutions before being labeled as such.

For the loss weights, we observe that the balance between reconstruction and perceptual loss depends on the image characteristics. When the data contain periodic patterns, higher weight on the reconstruction loss is effective, as the pixel-wise $L_2$ norm captures patch-level similarities well. In contrast, for images with stochastic structures where regularities manifest as textures, emphasizing the perceptual loss provides better alignment with the shared patterns. In Figure \ref{fig:qualitative_grid}, the first column (grid) illustrates a periodic pattern, while the second column (tile) demonstrates a stochastic pattern.
\subsection{Identity Mapping}
To prevent the model from collapsing into an identity mapping, we introduce several regularization techniques. First, patch permutation can be applied, where outputs are randomly shuffled so that the model is forced to reconstruct different patches. Second, random masking of image pixels encourages the network to capture underlying structures rather than simply replicating the input. Finally, in our experiments, adding Gaussian noise with mean zero and standard deviation 0.01 to the image proved particularly effective in preventing trivial solutions.
\subsection{Anomaly Score Computation}
In the inference stage, the anomaly heatmap is computed using Equation \ref{eq:anomaly_score}:

\begin{align}
\label{eq:anomaly_score}
&S_y = \Lambda_{r=1}^R \;\Bigg(\alpha_{rec}\mathcal{N}_{\min\max}\big(\frac{1}{N_r}\sum_{p \in \mathcal{P}_r} \;\!\,\\&R_{p,r} \odot\|R_{p,r} \odot  y - R_{p,r} \odot f_\theta(y)\|_2^2 \ \big)\notag -\alpha_{perc}\mathcal{N}_{\min\max}\;\!\\&\Big(\,\frac{1}{N_r}\sum_{p \in \mathcal{P}_r}R_{p,r} \odot\langle \phi(R_{p,r} \odot y), \phi(R_{p,r} \odot f_\theta(y)) \rangle\,\Big)\Bigg)\notag
\end{align}
where $\alpha_{rec}$ and $\alpha_{perc}$ are the weighting coefficients for reconstruction and perceptual loss, constrained such that $\alpha_{rec} + \alpha_{perc} = 1$. $\mathcal{N}_{\min\max}(M) = \frac{M - \min(M)}{\max(M) - \min(M)}$,
where $\min(M)$ and $\max(M)$ denote the minimum and maximum entries of the matrix $M$, respectively.
\section{Experiments}
\label{sec:exp}
\subsection{Datasets}
We evaluate our method on two standard benchmark datasets: MVTec-AD \cite{bergmann2019mvtec} and the HKBU fabric dataset \cite{6803963_HKBU_fabric}. 
MVTec-AD is a high-resolution benchmark for surface defect detection. 
We focus on the \emph{Grid} category, which contains 57 defective images with 5 defect types and 170 annotated defective regions. 
For the HKBU dataset, we use the dot-patterned fabric subset, which exhibits stochastic textures. 
This subset includes 30 defective images across 6 defect types, with 5 images per defect.
\subsection{Implementation Details}
We resize each test image to $256 \times 256$. For the perceptual loss, we use features from the eighth layer of VGG19 \cite{simonyan2014very_VGG}. To ensure compatibility with the VGG input, each patch is resized to $224 \times 224$. In Equation~\ref{eq:optimization}, we use a single patch size of $16 \times 16$. 

For the \emph{Grid} dataset, we train for up to 10 epochs with a stopping threshold of $10^{-4}$ and set $(w_{rec}, w_{perc}) = (1,0)$. 
For the \emph{fabric} dataset, we train for 1 epoch with a stopping threshold of $-100$ and set $(w_{rec}, w_{perc}) = (0,1)$. 

For anomaly score computation, we use $(\alpha_{rec}, \alpha_{perc}) = (1,0)$ across all experiments, except for the \emph{knots} anomaly type in the dot-patterned fabric dataset, where we use $(\alpha_{rec}, \alpha_{perc}) = (0,1)$.

For the ablation studies, we use the Grid dataset by selecting one image from each anomaly category, resulting in a total of five images. 
For the noise-level ablation, we add Gaussian noise with zero mean and a standard deviation varied between 0 and 0.1. 
For the masking ablation, we randomly mask out a varying proportion of pixels, with the masking ratio ranging from 0\% to 10\%. 
\subsection{Benchmarks and Metrics}
We compare our method against four state-of-the-art approaches: 
PG-LSR \cite{cao2017fabric_PGLSR}, a guided low-rank decomposition method; 
SR \cite{hou2007saliency_SR}, a spectral residual approach; 
GLCM \cite{raheja2013real_GLCM,haralick1973textural-GLCM_orig}, a statistical texture descriptor based on gray-level co-occurrence matrices; 
and WinCLIP \cite{jeong2023winclip}, which leverages CLIP \cite{radford2021learning_CLIP} by sliding a window over the image and computing similarity scores between patches and prompt templates to distinguish anomalies from normal regions.

For evaluation, we use the Area Under the Receiver Operating Characteristic curve (AUROC) and the Area Under the Precision–Recall Curve (AUPRC), both standard metrics in anomaly segmentation. 
AUROC measures the model’s ability to separate normal and anomalous pixels across thresholds by comparing true positive and false positive rates. 
AUPRC captures the trade-off between precision and recall over different thresholds.
\subsection{Anomaly Segmentation Comparison}
For both datasets, our method outperforms all competing approaches by a large margin. On the fabric dataset, SSDnet achieves an AUROC of 0.98 and AUPRC of 0.67, surpassing the second-best method (PG-LSR) by 0.02 and 0.10, respectively. Notably, WinCLIP—which builds on CLIP pretrained on OpenAI’s private WebImageText (WIT) dataset of 400 million image–text pairs—performs poorly, with only 0.80 AUROC and 0.21 AUPRC.

On the MVTec-AD Grid category, SSDnet achieves 0.99 AUROC and 0.60 AUPRC, substantially outperforming SR, the second-best method, which attains 0.97 AUROC and 0.45 AUPRC.
\begin{table*}[t]
\centering
\caption{Results on dot-patterned fabric dataset and MVTec-AD grid category dataset.}
\label{tab:wacv_dot_grid_results}
\setlength{\tabcolsep}{8pt}
\begin{tabular}{l l c c c c c}
\toprule
Dataset (support) & Metric $\uparrow$ & PG\text{-}LSR & GLCM & WinCLIP & SR & \textbf{SSDnet} \\
\midrule
\multirow{2}{*}{Fabric (30)} 
   & AUPRC $\uparrow$ & 0.57 & 0.38 & 0.21 & 0.19 & \textbf{0.67} \\
   & AUROC $\uparrow$ & 0.96 & 0.65 & 0.80 & 0.75 & \textbf{0.98} \\
\midrule
\multirow{2}{*}{MVTec-AD grid (57)} 
   & AUPRC $\uparrow$ & 0.34 & 0.02 & 0.21 & 0.45 & \textbf{0.60} \\
   & AUROC $\uparrow$ & 0.92 & 0.47 & 0.88 & 0.97 & \textbf{0.99} \\
\bottomrule
\end{tabular}
\end{table*}

\begin{figure}[t]
    \centering
    \includegraphics[width=1\linewidth]{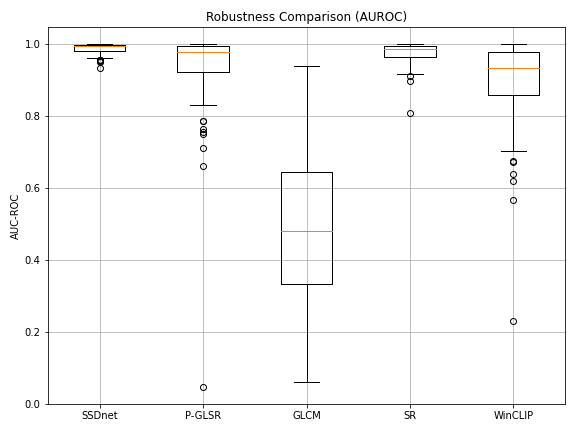}
    \caption{Box plots of AUROC.}
    \label{fig:box_roc}
\end{figure}

\begin{figure}[t]
    \centering
    \includegraphics[width=1\linewidth]{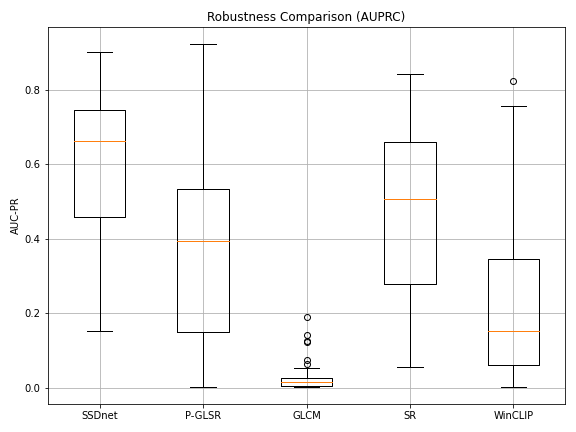}
    \caption{Box plots of AUPRC.}
    \label{fig:box_prc}
\end{figure}

Beyond mean values, we also analyze per-image variability using box plots of AUROC and AUPRC on the Grid dataset. For AUROC (Figure~\ref{fig:box_roc}), SSDnet achieves the highest median and the lowest inter-quartile range, indicating highly consistent performance. For AUPRC (Figure~\ref{fig:box_prc}), SSDnet again has the highest median, with a spread comparable to WinCLIP but much smaller than SR and PG-LSR. Although GLCM shows the lowest spread, its performance is poor, with only 0.02 AUPRC.
\section{Ablation Studies}
\label{sec:ablation}
\subsection{Performance With Noisy Data}
\begin{figure}[t]
    \centering
    \includegraphics[width=1\linewidth]{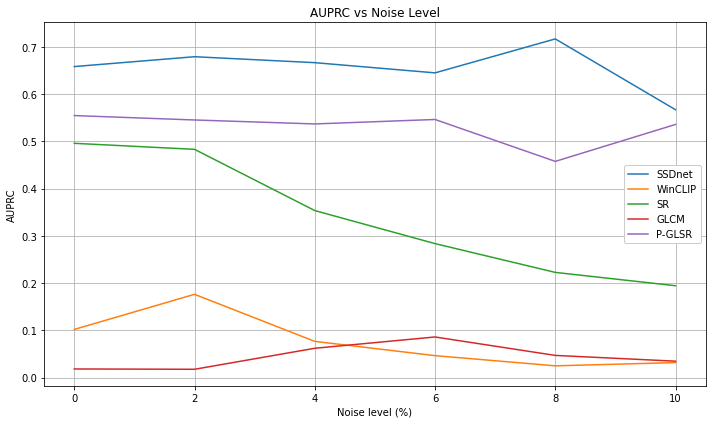}
    \caption{AUPRC vs Noise Level}
    \label{fig:auprc_noise}
\end{figure}

\begin{figure}[t]
    \centering
    \includegraphics[width=1\linewidth]{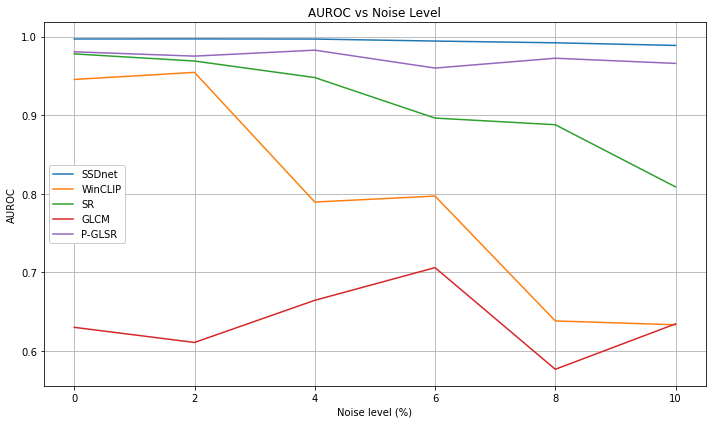}
    \caption{AUROC vs Noise Level}
    \label{fig:auroc_noise}
\end{figure}

To assess robustness to noise, we evaluate SSDnet and competing methods under increasing noise levels. As shown in Figures~\ref{fig:auprc_noise} and \ref{fig:auroc_noise}, SSDnet consistently outperforms all baselines across all noise levels. Notably, Spectral Residual (SR) and WinCLIP degrade sharply as noise increases, while PG-LSR remains relatively stable, similar to our method.
\subsection{Performance With Missing Pixels}
When an image contains missing values, we apply SSDnet in two stages. First, we use it to inpaint the missing pixels and obtain a reconstructed image. Then, we apply SSDnet again on the reconstructed image to detect anomalies. 

\begin{figure}[t]
    \centering
    \includegraphics[width=1\linewidth]{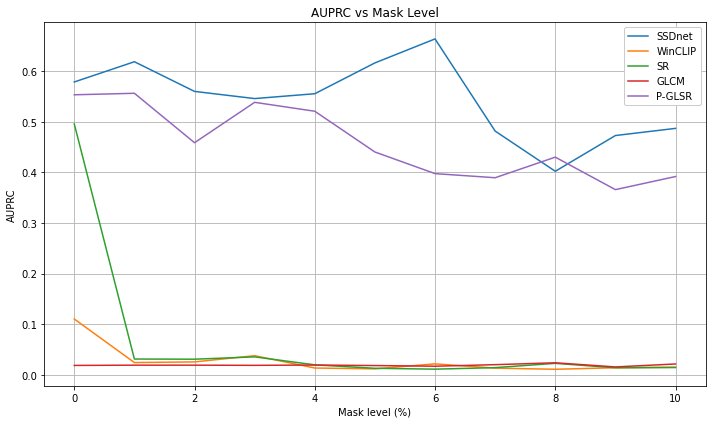}
    \caption{AUPRC vs Mask Level}
    \label{fig:auprc_mask}
\end{figure}

\begin{figure}[t]
    \centering
    \includegraphics[width=1\linewidth]{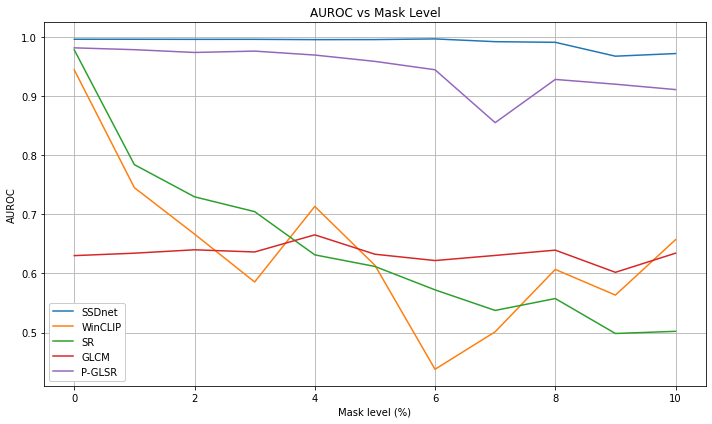}
    \caption{AUROC vs Mask Level}
    \label{fig:auroc_mask}
\end{figure}

For masking, SR and WinCLIP are highly sensitive: even 1\% random masking causes their performance to drop near zero, as shown in Figure~\ref{fig:auprc_mask}. In contrast, GLCM remains unaffected, maintaining nearly constant AUROC across masking levels (Figure~\ref{fig:auroc_mask}). Among the stronger methods, SSDnet and PG-LSR achieve the best results, with SSDnet generally outperforming PG-LSR across masking levels and metrics, except at 8\% masking where PG-LSR holds a slightly higher AUPRC.
\subsection{Additional Qualitative Results}
\begin{sidewaysfigure}
  \centering
  \resizebox{0.8\textwidth}{!}{
    \begin{tabular}{@{}ccccccc@{}}
      \textbf{Anomalous} & \textbf{SSDnet (heatmap)} & \textbf{SSDnet (mask)} &
      \textbf{GLCM} & \textbf{PG-LSR} & \textbf{SR} & \textbf{WinCLIP} \\
      \midrule
      \qualrow{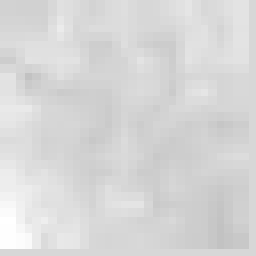}
      \qualrow{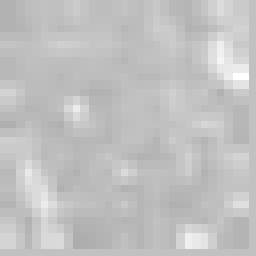}
      \qualrow{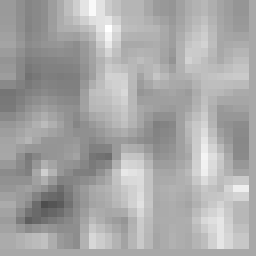}
    \end{tabular}
  }
  \caption{Qualitative single-image anomaly segmentation results. Each column corresponds to one test image; columns (left→right) show the grid, tile and wood categories; rows (bottom→top) show the anomalous input, our method’s anomaly heatmap and binary prediction mask, and baselines: GLCM energy, PG-LSR, SR saliency, and WinCLIP heatmap. Our patch-based, training-free approach reconstructs the normal pattern and highlights deviations as anomalies.}
  \label{fig:qualitative_grid}
\end{sidewaysfigure}

We further evaluate SSDnet qualitatively on additional MVTec-AD categories, including grid, tile, and wood (Figure~\ref{fig:qualitative_grid}). From bottom to top, each row shows the original image, SSDnet heatmap, SSDnet binary map (via Otsu’s thresholding \cite{4310076_OTSU}), and heatmaps from GLCM, PG-LSR, SR, and WinCLIP. While PG-LSR produces reasonable results, its anomaly maps often overestimate defect regions or miss anomalies entirely, as in the right column where one of three connected components is undetected. Additional qualitative results on carpet and leather categories can be found in Figure \ref{fig:qualitative_grid_appendix} in Appendix \ref{app:add_qual}.

\section{Conclusion}
In this paper, we introduced SSDnet, a multi-resolution patch-based model for pixel-level anomaly segmentation in single images, without requiring any training data. We demonstrated that SSDnet outperforms state-of-the-art approaches, including foundation model–based methods, spectral residual methods, low-rank decomposition techniques, and statistical descriptors. We further evaluated its robustness to noise and masking, showing that SSDnet remains largely unaffected by both. Finally, we employed a perceptual loss based on the unnormalized inner-product of embeddings, enabling the model to better capture stochastic textures and complex image structures.

Although our method demonstrates excellent performance, its main limitation lies in the need to fine-tune parameters for each specific domain. This limitation could be addressed in a multi-shot setting, where the model can be fine-tuned on a small set of samples using cross-validation. However, in the zero-shot setting, such adaptation is not feasible, and careful manual selection of parameters becomes necessary. 

For future research, this method could be extended to anomaly detection in unstructured point cloud data, as well as to other modalities such as non-stationary time series.
\label{sec:conclusion}
\clearpage
{\small
\bibliographystyle{ieee_fullname}
\bibliography{egbib}
}
\newpage
\section{Appendix}
\subsection{Perceptual Loss}
\label{app:perceptual}
In this section, we prove that cosine similarity between embeddings is equivalent to normalized Euclidean distance. Let two embeddings be
$u = \frac{f(y_1)}{\|f(y_1)\|_2}, \quad v = \frac{f(y_2)}{\|f(y_2)\|_2}$,
where $y_1, y_2$ are two different image patches.
The normalized Euclidean distance is:
$\|u - v\|_2^2 = \|u\|_2^2 + \|v\|_2^2 - 2\langle u,v \rangle = 2 - 2\langle u,v \rangle.$
Since $\langle u,v \rangle = \cos(f(y_1), f(y_2))$ we obtain $
\|u - v\|_2^2 = 2 - 2\cos(f(y_1), f(y_2)).$
Thus, minimizing normalized Euclidean distance is equivalent to maximizing cosine similarity, up to an additive constant.

In contrast, the perceptual loss we employ is defined as
$\langle f(y_1), f(y_2) \rangle = \|f(y_1)\|_2 \|f(y_2)\|_2 \cos(f(y_1), f(y_2))$
which enforces both angle alignment (as in cosine similarity) and the growth of embedding magnitudes. This enforces discriminative features to dominate the optimization, thereby facilitating more accurate learning of normal patterns.

\subsection{Additional Qualitative Results}
\label{app:add_qual}
\begin{figure*}[t]
  \centering
  \setlength{\tabcolsep}{4pt} 
  \renewcommand{\arraystretch}{1.0}
  \begin{tabular}{@{}ccccccc@{}}
    \textbf{Anomalous} & \textbf{SSDnet (heatmap)} & \textbf{SSDnet (mask)} &
    \textbf{GLCM} & \textbf{PG-LSR} & \textbf{SR} & \textbf{WinCLIP} \\
    \midrule
    \qualrow{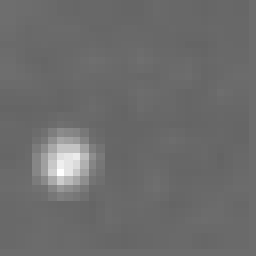}
    \qualrow{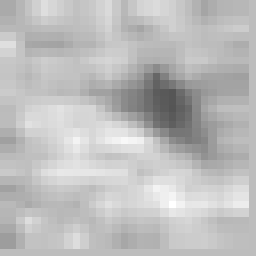}
    \qualrow{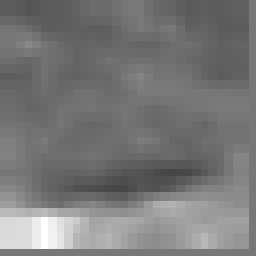}
  \end{tabular}
  \caption{Additional qualitative single-image anomaly segmentation results. 
  Each row corresponds to one test image; columns ((bottom→top) show leather, leather, and carpet categories; rows (left→right) show the anomalous input, our method’s anomaly heatmap and binary prediction mask, and baselines: GLCM energy, PG-LSR, SR saliency, and WinCLIP heatmap. 
  Our patch-based, training-free approach reconstructs the normal pattern and highlights deviations as anomalies.}
  \label{fig:qualitative_grid_appendix}
\end{figure*}
\end{document}